\documentclass[10 pt, conference]{ieeeconf}

\IEEEoverridecommandlockouts                              

\usepackage{xcolor}

\definecolor{MyDarkBlue}{rgb}{0,0.08,1}
\definecolor{MyDarkGreen}{rgb}{0.02,0.6,0.02}
\definecolor{MyDarkRed}{rgb}{0.8,0.02,0.02}
\definecolor{MyDarkOrange}{rgb}{0.40,0.2,0.02}
\definecolor{MyPurple}{rgb}{111,0,255}
\definecolor{MyRed}{rgb}{1.0,0.0,0.0}
\definecolor{MyGold}{rgb}{0.9,0.8,0.0}
\definecolor{MyDarkgray}{rgb}{0.66, 0.66, 0.66}

\usepackage{graphics} 
\usepackage{epsfig} 
\usepackage{mathptmx} 
\usepackage{times} 
\usepackage{amsmath} 
\usepackage{amssymb}  
\usepackage{color}
\usepackage{gensymb}
\usepackage{siunitx}

\usepackage[noadjust]{cite}

\newcommand{\myparagraph}[1]{\vspace{0.1in}\noindent\textbf{#1}}

\title{\LARGE \bf A Passively Bendable, Compliant Tactile Palm with RObotic Modular Endoskeleton Optical (ROMEO) Fingers

}

\author{
    \authorblockN{Sandra Q. Liu and Edward H. Adelson} 
        \authorblockA{Massachusetts Institute of Technology\\
    {\tt\small sqliu@mit.edu, adelson@csail.mit.edu}
    } } 

\usepackage{multirow}
\usepackage{comment}

\begin{document}

\maketitle
\thispagestyle{empty}
\pagestyle{empty}

\begin{abstract}

Many robotic hands currently rely on extremely dexterous robotic fingers and a thumb joint to envelop themselves around an object. Few hands focus on the palm even though human hands greatly benefit from their central fold and soft surface. As such, we develop a novel structurally compliant soft palm, which enables more surface area contact for the objects that are pressed into it. Moreover, this design, along with the development of a new low-cost, flexible illumination system, is able to incorporate a high-resolution tactile sensing system inspired by the GelSight sensors. Concurrently, we design RObotic Modular Endoskeleton Optical (ROMEO) fingers, which are underactuated two-segment soft fingers that are able to house the new illumination system, and we integrate them into these various palm configurations. The resulting robotic hand is slightly bigger than a baseball and represents one of the first soft robotic hands with actuated fingers and a passively compliant palm, all of which have high-resolution tactile sensing. This design also potentially helps researchers discover and explore more soft-rigid tactile robotic hand designs with greater capabilities in the future. 

\end{abstract}

\section{Introduction}

When humans power grasp an object, they envelop their whole hand around the object. Not only do the fingers curve around the object surface, but the palm does also. In particular, the palm has a 1-DOF (degree of freedom) active joint at the base of the four, non-thumb metacarpal bones. This joint, along with the compliance of human skin, allows human hands to fully wrap itself around various sized objects. Moreover, from a sensing perspective, the palm can provide sensory information that the fingers or fingertips alone cannot provide.

However, many existing robotic hands focus on the development of fingers, and the palm is an afterthought. Very dexterous fingers or ones with a very large bending radius can already slightly emulate this ``wrapping’’ behavior of which the palm is capable \cite{Liu2023GelSightEA}. Other hand designs experiment with the thumb abduction in the palm to replicate an enveloping grasp \cite{parkhrihand, 26DOF}. Still, others utilize a compliant and soft material as a palm to help passively enfold the object being grasped instead of focusing on the palmar abduction joint \cite{changeablepalm}; the softness also serves as a safe way to interact with the world and other people \cite{wang2018toward}.

\begin{figure}[ht!]
    \centering
    \includegraphics[width=1.0 \linewidth]{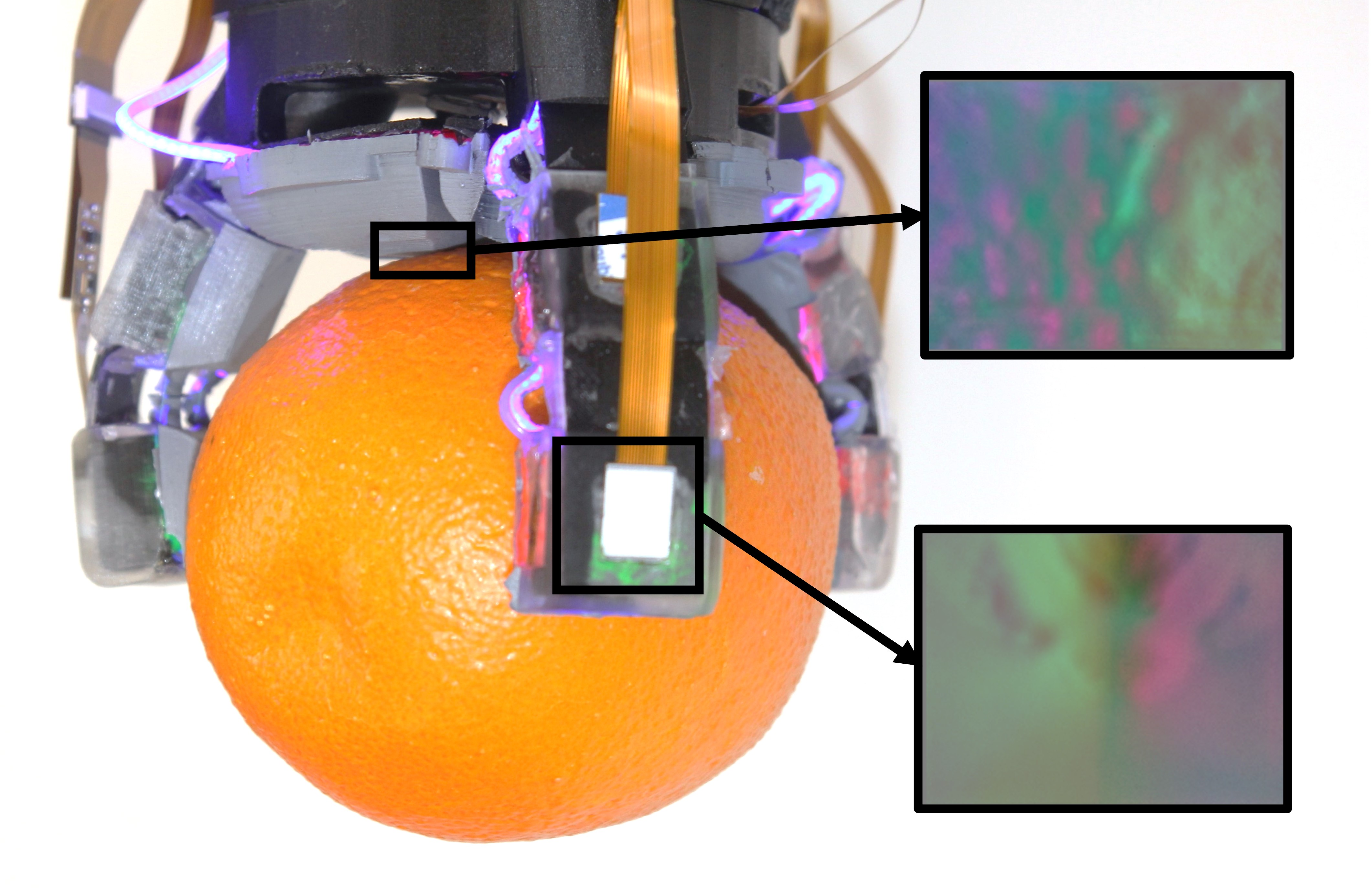}
    \caption{A novel dual compliant palm with three novel ROMEO fingers in a 120$^{\circ}$ configuration grasping an orange. Both the palm and the fingers have GelSight-inspired tactile sensing through usage of the newly designed flexible illumination system. To the right of the image are the resulting difference images of the bumpy orange skin sensed by the palm and the orange stem sensed by the ROMEO finger tip.}
    \label{fig:teaser}
    \vspace{-12pt}
\end{figure}

In general, we want a palm-like structure that is able to comply to and encase the objects that are being grasped. We can do this using a soft material atop a passive, compliant structure, which emulates a 1-DOF joint. This compliant palm, along with tactile, soft-rigid fingers, will be able to help the resulting hand perform a better enveloping grasp than if the palm were not compliant. Moreover, palms can benefit from tactile sensing, which will provide more information for object or contact classification. Finally, to enable ease of use and manufacturing, we want to simplify the hardware components of the cost-effective camera-based tactile sensing we use and also modularize the fingers. 

We present the following contributions:

\begin{itemize}
    \item The development of a new low-cost, effective illumination system for camera-based tactile sensors that can be easily integrated into different soft bodies.
    \item A RObotic Modular Endoskeleton Optical (ROMEO) finger design, which uses the new illumination strategy.
    \item The novel designs of passively bendable and compliant palms with high-resolution tactile sensing that are capable of increasing the surface area contact of a grasped object, which can be incorporated into a fully high-resolution tactile sensing hand (Fig. \ref{fig:teaser}).
\end{itemize}

\section{Related Work}

\subsection{Robotic Hands and Palms}

Most existing robotic hands focus on the development of the fingers, from rigid to soft to a hybrid combination of both. In papers where a palm is developed, most researchers look at the thumb abduction to help create a cupping motion for grasping. For example, Park and Kim develop an open-source anthropomorphic rigid robotic hand with a simple 1-DOF thumb abduction joint \cite{parkhrihand}. Tangentially, Xiong et al. design a more complicated rigid thumb joint with a helical gear set to allow it to perform better grasps on different objects \cite{xiong2016rigid}. These rigid robotic hand designs, with the addition of the thumb joint in the palm, allow the hands to grasp many different types of objects. Similarly, many soft hands approach palmar actuation by solely focusing on the thumb abduction. Both Fras and Althoefer, and Zhou et al. design a thumb abduction joint for the palm using soft rubber materials, which can help conform better to objects than their rigid material counterparts \cite{biomimeticprosthetic, thumbabduct}. For the novel soft-rigid hybrid gripper developed by Tavakoli and de Almeida \cite{tavakoli2014flexture1}, the only palmar bending comes from the thumb joint bending towards the direction of the other fingers. For the most part, a common theme of many hand design papers are the lack of this 1-DOF actuation in the center of the palm.

Nonetheless, there are a few papers that explore palm concavity and changeable palms. Pagoli et al. develop a linearly actuated palm that can extend to engage with an object using suction or vacuum jamming \cite{pagoli_vacuumpalm}, while Sun et al. design a changeable palm which allows the fingers to rotate relative to the central palm so that the fingers can more easily optimize a grasping pose \cite{changeablepalm}. From a more human-inspired direction, Wang et al. create a fully pneumatic soft hand with a palm that is able to perform both palmar bending and thumb abduction \cite{humansoftpalm}. In a somewhat similar vein, Li et al. fabricate a passive rotational palm joint with a jamming material palm that can engage when the fingers press an object into the palm \cite{li_vacuumpalm}. Capsi-Morales et al. also design and manufacture a novel, rigid human-inspired hand that is capable of curling the palm length-wise and can simulate a cupping grasp of the hand \cite{coolpaint}. This hand cupping motion is also replicated in a pneumatic soft hand developed by Zhou et al., which has 26 actuated degrees of freedom \cite{26DOF}. 

All of these papers show that the inclusion of compliance, curvature, or a changeable element in the palm helps to achieve better and more stable grasps. However, they all require additional actuation, which can potentially make them difficult to incorporate in other hand designs or to implement for different hand configurations. They could also all be improved with the addition of tactile sensing so that they can perform more identification or contact classification tasks.

\subsection{Tactile Sensing for Robotic Hands}

Although robotic hands have achieved a great variety of tasks using only external vision \cite{visualpeginhole, visualcable}, there are situations where such cameras will not be as useful, such as in a highly-occluded environment or digging through clutter. In these scenarios, having tactile sensing is necessary. 

A wide variety of tactile sensors for soft fingers and hands exists, ranging from piezoresistive to capacitive to optical wave guides to pressure and strain sensors \cite{shadowpalmsense, CapacitiveSkinFingers, fiberoptic, shorthose_softhandsense, softstrainreview}. Lei et al. also fabricate a biomimetic tactile palm which is composed of an array of sensing electrodes and a conductive liquid inside of a compliant skin \cite{biotactilepalm}. However, many of these sensors are low-resolution and potentially do not have the capability to distinguish or classify an object with one touch and instead require multiple touches \cite{palm_sense}.

One method for achieving high-resolution tactile sensing is to use camera-based sensors, like the GelSight-inspired ones, which are simple, low-cost, and effective \cite{dong2017improved, OGGelSight}. For these sensors, a camera looks at an illuminated piece of elastomer atop a rigid structure, and the camera can track the touch information on the illuminated surface. The simple concept of these sensors allows for them to exist in multiple form factors, such as wedge shapes, round finger-like domes, and curved claw-like structures \cite{wedge, 360, digit, curve}. Even though most of these sensors require a rigid backing to work, recent developments have been made to integrate camera-based tactile sensors into soft, compliant robots \cite{ExoGelSight, LiuFinRay, bbfinray}. In particular, there was the development of a soft-rigid finger with continuous high-resolution tactile sensing along its front and sides \cite{Liu2023GelSightEA}. Three of these fingers were placed into a gripper and could classify objects using a singular grasp due to the rich tactile information provided. However, the addition of a palm would perhaps greatly improve its classification results and grasp stability, which it struggled with for certain objects. To this end, we design this dual-compliance high-resolution tactile sensing palm, which is then integrated into a fully tactile hand. 

\section{Methods}

\subsection{Hardware}

\begin{figure}[ht!]
    \centering
    \includegraphics[width=0.85 \linewidth]{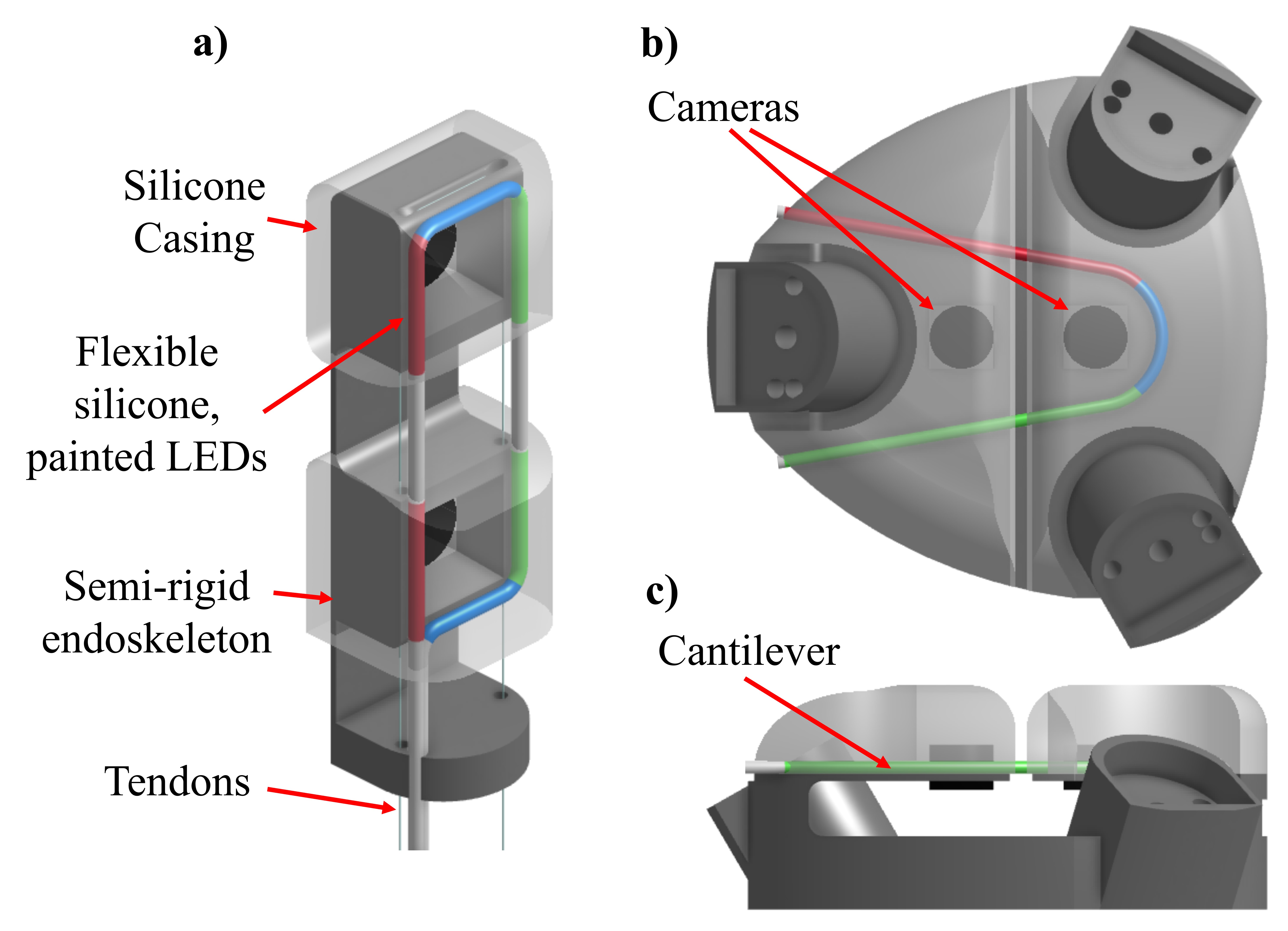}
    \caption{Rendered CAD files of the ROMEO finger (a) and the compliant palm (b, c), showing the various labeled parts.}
    \label{fig:palmfinger}
    \vspace{-6pt}
\end{figure}

We develop a new camera-based tactile sensing illumination system and a novel tactile, compliant palm, which can be used in a variety of different configurations with our ROMEO fingers. These designs are shown in Fig. \ref{fig:palmfinger}.

\myparagraph{Illumination}

Most GelSight-inspired sensors consist of a camera viewing an elastomer illuminated by directional red, green, and blue (RGB) LEDs. This concept takes advantage of the RGB color channels in a single camera image, meaning that a single image of a tactile indentation allows the sensor to perform 3D reconstruction using the color gradients. We wish to emulate this illumination method so that it can more easily be applied in soft robots.  

One of the issues with electronic integration into silicone soft robotic bodies is the delamination that occurs when the silicone deforms around their encased rigid electronics. This delamination creates an air interface between the LEDs and the silicone, which can cause unpredictable light hot spots and a lack of blended light gradients on the surface of the gel elastomer.

To mitigate this issue and to simplify the electronics, we turn to commercial flexible 1.7 mm-diameter LED filaments (Adafruit), which are easy to use and enclosed in silicone. The latter makes these strands easily to integrate into silicone and the former enables them to flex with different soft bodies. Their only downside is that the LED colors in one strand are not customizable, making them difficult to use for different lighting configurations. 

We turn to inspiration from the synthesized fluorescent silicone paint formulated for the Baby GelSight Fin Ray \cite{bbfinray}, with the idea that if a section of fluorescent paint covers a section of the blue LED strand, that section will fluoresce the color we choose. Similarly, we can apply this concept to a white light LED strand and use semi-transparent silicone paints. Furthermore, because silicone bonds to itself, the paint will stay bonded to the LEDs, and it will also stay bonded to the silicone gel body, preventing delamination.

The fluorescent paint, which follows the formula from \cite{bbfinray} and uses acrylic paint (Liquitex) and silicone adhesive, is applied to the visible sides of the flexible LED filament after it is adhered to a surface. To get good coloration, we use a Q-tip to apply at least a 1 mm surface of paint to all visible sides so that the paint covers the blue LEDs.

On the other hand, for the white LEDs, we want to ensure that the paint is semi-transparent so that the light can shine through the paint. As a result, we utilize Silc Pig pigments (Smooth-On Inc), which are more translucent. Our formulation is 10 parts of silicone adhesive mixed with 1 part of the red, green, or blue silicone pigments, and then diluted with 3 parts NOVOCS Gloss (Smooth-On Inc). We apply at most a 1 mm layer of paint. 

This technique allows us to more easily use flexible LED filaments for different configurations without the need to design a new PCB for every part or component of our soft robot bodies. The materials also bond easily to one another and can prevent delamination. We use these painted flexible LEDs in both the ROMEO finger and compliant, tactile palm designs. 

\myparagraph{ROMEO Finger Design}

We utilize a soft-rigid structure inspired by the bone and flesh structure of a human hand. Doing so allows us to leverage the strength provided by the structural rigidity and also allows us to take advantage of the soft compliance of the gel. The inner structure is dubbed an ``endoskeleton'' and is 3D printed out of nylon and chopped carbon fiber material (Onyx, Markforged). Using Onyx allows us to have a compliant hinge, which can easily be actuated using tendons, and its carbon fiber content gives the endoskeleton structure rigidity. Additionally, the endoskeleton is able to house both the camera and LED structure, similar to how it was for \cite{Liu2023GelSightEA}.

Once the structure is 3D printed, Onyx braided line (Piscifun) is threaded through 3D printed holes at the top surface of the endoskeleton. Afterwards, a flexible LED filament is placed on top of the front and adhered using a 3D printed PLA guide and a silicone adhesive. Once the silicone adhesive is cured, the guide is removed from the structure. 

Depending on which colored LED was incorporated, we mix the corresponding silicone paints and apply them to the appropriate sections of the flexible LED filament. In particular, each section of the filament surrounding the camera holder has to exhibit three different colors: red, green, and blue. 

After the LED filament is painted, the surface of the finger is coated in silicone adhesive to help prevent delamination of the silicone from the endoskeleton. Next, we prepare the mold for the silicone bodies that encase the finger. Although we do not polish the mold, we make sure that the parts of the mold for the camera mounts are covered with a single laser cut piece of thin acrylic, so that the viewing ports are clear. A paint formulation of 1 part catalyst to 10 parts gray silicone pigment (Raw Materials Inc.) to 1.25 parts 4 $\mu$m aluminum cornflakes (Schlenk) to 30 parts NOVOCS Gloss (Smooth-On Inc) is used for the tactile skin of the finger. This mixture is then blended using an ultrasonicator to ensure even distribution of aluminum flakes and gray pigment. The paint mixture is then airbrushed onto the top halves of the 3D printed PLA molds, which have been treated with Inhibit-X (Smooth-On Inc) and Ease Release 200 (Smooth-On Inc) for easy removal of the silicone. 

\begin{figure}[ht!]
    \centering
    \includegraphics[width=1.0 \linewidth]{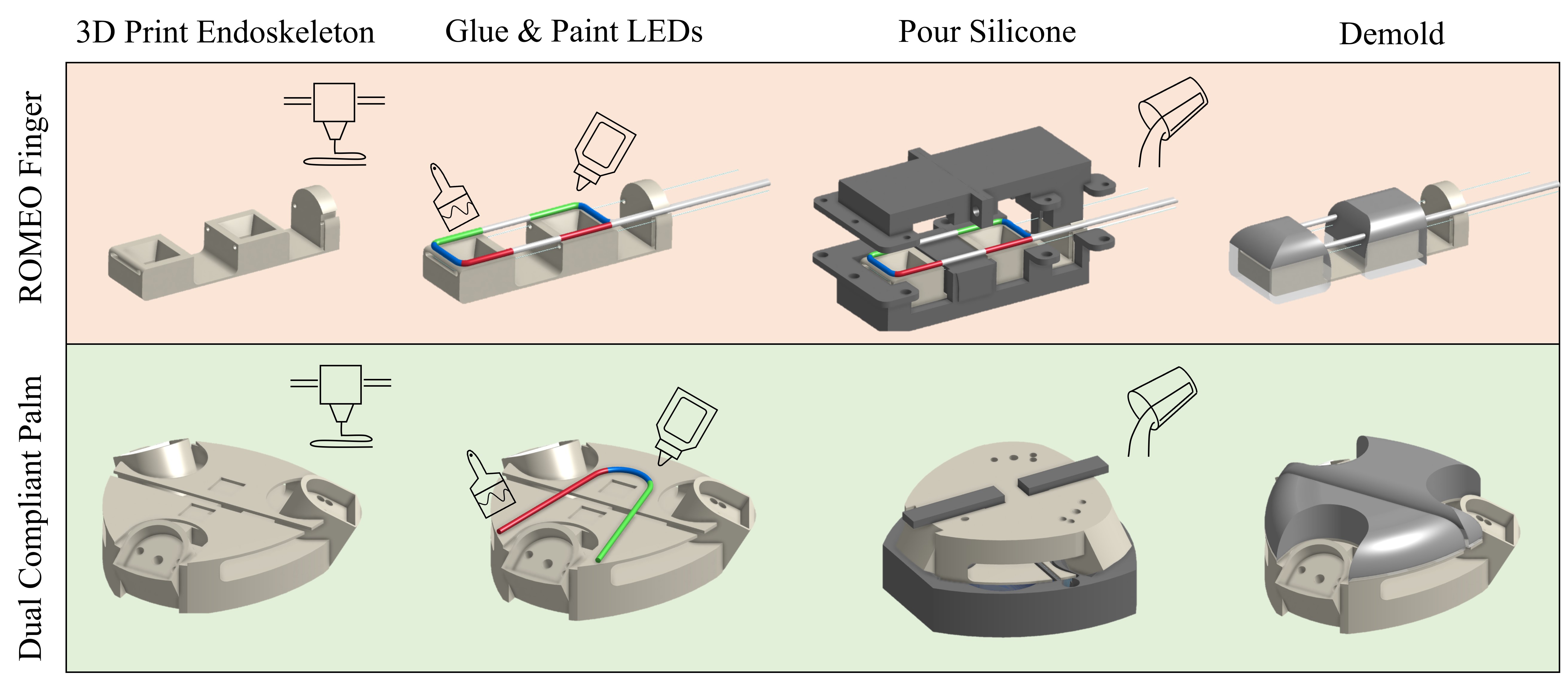}
    \caption{The manufacturing process of both the finger and the palm, which involve 3D printing the endoskeleton, prepping the flexible LED filaments, painting and pouring the silicone, and finally demolding the parts. Not pictured is the step where the aluminum paint mixture is airbrushed onto the mold before silicone is poured.}
    \label{fig:manu}
    \vspace{-12pt}
\end{figure}

We use XP-565 silicone (Silicones Inc.) because it is transparent, allowing our camera to see through the entire gel structure. The mixture we use is 1 part silicone catalyst to 15 parts silicone to 3 parts plasticizer (Phenyl Trimethicone, Lotioncrafter), which results in a softer gel. Before pouring the silicone mixture into the prepared molds, we spread silicone adhesive in between the various parts and screw the molds together to ensure that the mixture will not leak. Once the mixture is poured inside, we leave the molds on a vibrating plate for at least an hour to allow trapped air bubbles to escape. These molds are then either left to air cure for 24 hours or they are placed into a dehydrator at 50$^{\circ}$C for 6 hours. The lower cure temperature is to ensure that none of the plastic elements reach their glass temperature and cause the mold to warp. The main parts of the manufacturing process are shown in Fig. \ref{fig:manu}.

The camera housings are designed such that when the finger bends at a 45$^{\circ}$ angle relative to itself, the two 120 field of view (FOV) cameras in each finger segment are able to see a continuous tactile display of the entire front and sides of the finger. We chose not to envelop the entire endoskeleton structure to allow for a larger range of bending in the finger. The silicone gel we use, although compliant, is unable to elastically compress on itself even if grooves are added in between the finger links. By removing the gel in between, we enable the finger to bend until the edges of the separate silicone pieces jam against one another. Moreover, the silicone skin will not wrinkle, as it would have if the endoskeleton was fully encased in silicone. 

\myparagraph{Compliant, Tactile Palm Design}

For the palm design, we chose to incorporate two different types of compliance: structural and material. This design choice was inspired by the human palm, which exhibits both palmar bending at the base of the metacarpals and has soft, compliant flesh on the outside. Having both compliance types allows us to easily envelop various objects with the hand. 

We considered having active actuation for the palm, but decided the electronics would become too bulky, especially if each attached finger also required separate actuation. To simplify electronics and keep the form factor of the palm small, we decided to incorporate a type of passive compliance. One solution that is easy to implement in different palm shapes is the cantilever, which will deform if force is applied along its length. 

In particular, if we place multiple cantilever beams with their ends all concentrated in one area of the palm, we can achieve palmar bending passively as long as an object is pressed into the palm. As a result, we choose to use two cantilevers for ease of camera-based tactile sensing integration, but acknowledge the fact that multiple cantilevers (n $\ge$ 2) can be used for symmetry. 

\begin{figure}[ht!]
    \centering
    \includegraphics[width=1.0 \linewidth]{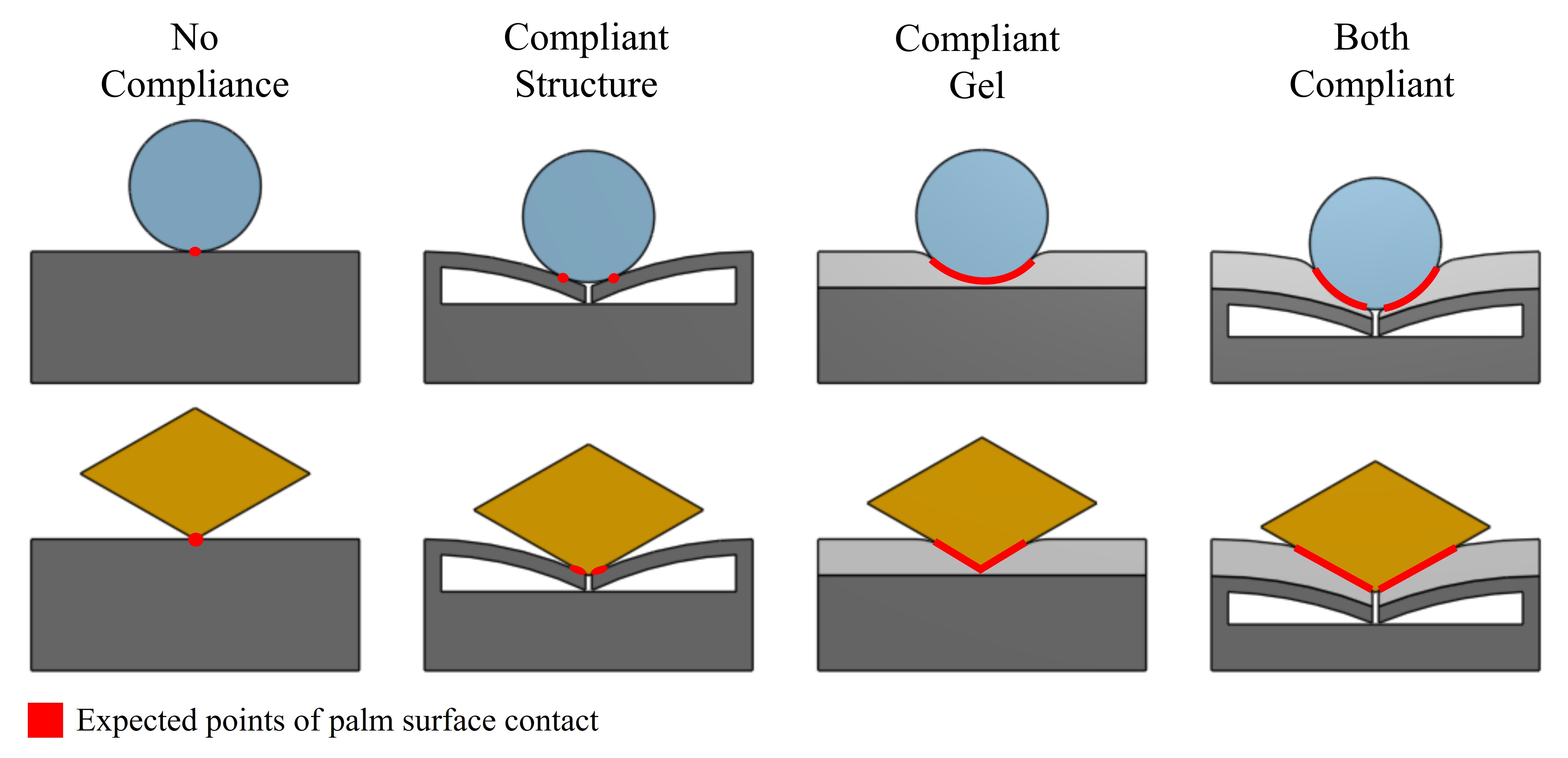}
    \caption{The expected points of contact of a circle and rhombus object into the different types of palms, ranging from no compliance to both structural and material compliance. We expect having both types of compliance to give us maximum surface area contact.}
    \label{fig:expected}
    \vspace{-10pt}
\end{figure}

This passive structural compliance, along with the compliance from the gel material enveloping the cantilever, will theoretically work better at enveloping an object than just implementing structural or material compliance by themselves (Fig. \ref{fig:expected}). For instance, if we take a cylindrical object and press it against a rigid surface, we expect there to be only one line of contact. However, if we integrate two cantilevers in the palm and press the same object, we will get at least two lines of contact, one per cantilever beam. The compliant gel will also provide multiple points of contact as it elastically compresses and conforms to the surface area of the object. Thus, when the two types of compliance are combined, we get the maximum surface area contact, which is better for a more stable grasp. Not only do we have the material gel compliance, we also get added surface area contacts when the cantilever beams deform and help to better wrap the gel surface around the area of the object.

To design such a cantilever beam so that it does not plastically deform, we assume a constant rectangular cross section as our baseline and use that to implement a conservative hard stop at the base of the palm. We let $F$ stand for force applied on the end of a cantilever of length $L$ and thickness $t$ with Young's modulus $E$ and moment of inertia $I$. Then, we use the following equations to determine the maximum deflection $\delta_{max}$ and stress $\sigma_{max}$:

\begin{equation}
    \delta_{max} = FL^3 / {3EI}
    \label{eq:delta}
\end{equation}

\begin{equation}
    \sigma_{max} = FLt / 2I
    \label{eq:sigma}
\end{equation}

Next, to find the conservative maximum deflection the beam can exhibit, we combine (\ref{eq:delta}) and (\ref{eq:sigma}) to get:

\begin{equation}
    \delta_{max} = \frac{2}{3}\sigma_{max}L^2 / Et
    \label{eq:calcdelta}
\end{equation}

We use the Onyx material properties as an estimate for $\sigma_{max}$ and $E$ values, making sure that we set $\sigma_{max}$ as half of the max tensile stress for the yield stress of the material and to account for a factor of safety. Our $t$ value is determined by the thinnest, structurally sound 3D printed beam we can obtain using the Markforged printer while $L$ is determined by the reach of the ROMEO fingers in their respective configurations. 

Once the $\delta_{max}$ of the beam has been calculated, the resulting geometry is designed and 3D printed using Onyx material, which can serve as a flexible hinge. Because the maximum deformation of the beam keeps the material in its elastic realm, we do not have to worry about plastic deformation over repeated, multiple use. 

After the palm base is printed, the flexible LED filament is adhered and painted. Most of the steps to creating the gel pad are similar to the steps for creating the ROMEO finger gels. The main difference is that the mold is a single part instead of three separate parts. We also add two 3D printed parts with an acrylic covering over the surface to the mold to ensure that there are camera holes in the palm. The camera is placed into the cantilever afterwards so that we can track the tactile deformation in the image without needing to separate out the palm structural deformation.

The thickness of the gel pad for the palm is partially determined using the camera FOV, so that the camera can see a relatively large portion of the palm center. One camera is placed in each cantilever, and they are able to view continuously along the central region of the palm. 

\myparagraph{Hand Structure}

\begin{figure}[ht!]
    \centering
    \includegraphics[width=1.0 \linewidth]{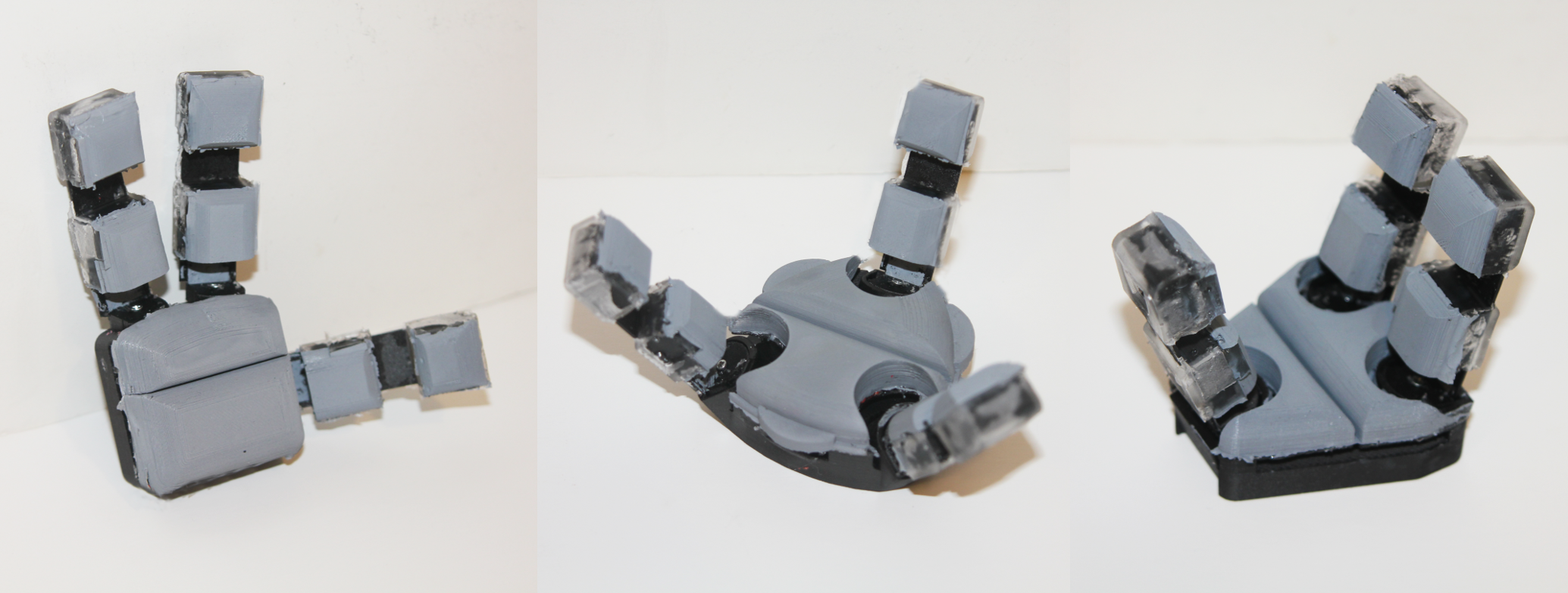}
    \caption{Our ROMEO fingers and three different passively double compliant palm configurations. From left to right, we have an anthropomorphic, a 120$^{\circ}$, and a ``Y'' configuration gripper.}
    \label{fig:modular}
    \vspace{-3pt}
\end{figure}

Each ROMEO finger has a base, which allows it to slot into different modular palm configurations. To showcase this modularity, we also design multiple palm configurations, all with both structural and material compliance (Fig. \ref{fig:modular}). We also note that although we do not show it in this paper, the designed hands can have fewer or more fingers.  

\begin{figure}[ht!]
    \centering
    \includegraphics[width=0.8 \linewidth]{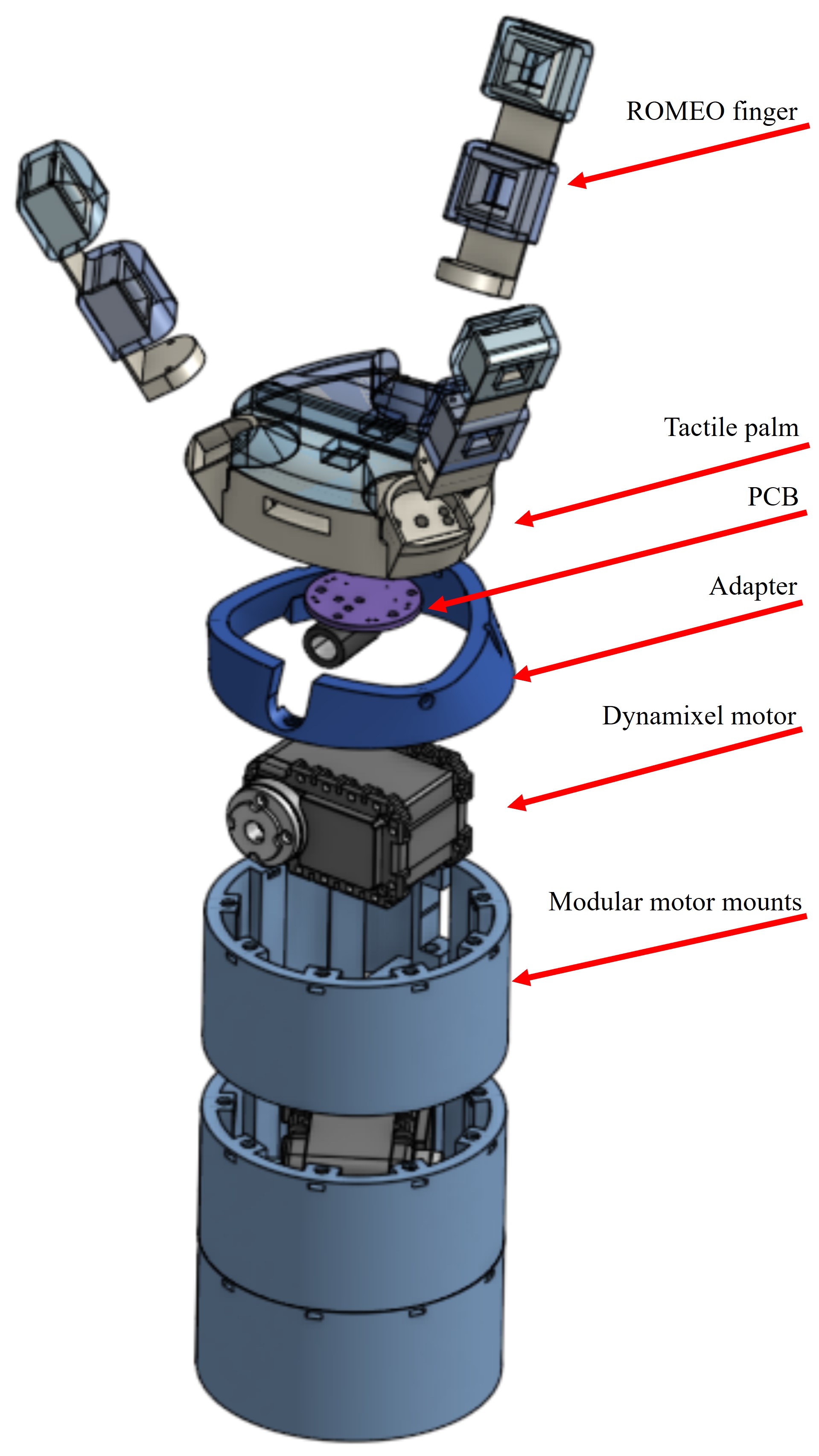}
    \caption{The exploded view of our entire hand structure, including the modular motor mounts and some electronics. Not pictured are the heated inserts, bolts, wires, cameras, tendons, and flexible LEDs.}
    \label{fig:explosion}
    \vspace{-8pt}
\end{figure}

After we slot the fingers into the palm, the palm is then attached to the electronic components and their housing, all of which are shown in Fig. \ref{fig:explosion}. Individual parts are connected to one another using M2 heated inserts.

The PCB connects each of the flexible LED filaments in parallel with a 3 V source. An adapter is custom designed for the palm so that it can fit atop a modular motor mount, which houses a Dynamixel AX-12A motor. We specifically design a modular motor mount so that it can be used for different hand configurations and for multiple fingers. The motor is attached to the base of each mount, which has rotationally symmetric mounting holes, allowing the mounts to be arranged in the proper orientation for arbitrary finger configurations. 

\subsection{Software}

The software components mainly involve the tactile sensing. We use 120$^{\circ}$ FOV Raspberry Pi spy cameras and connect each of them to a separate Raspberry Pi board. The camera images are streamed using mjpg-streamer and processed with OpenCV and the fast Poisson solver \cite{opencv, poisson}. 

\begin{figure}[ht!]
    \centering
    \includegraphics[width=1.0 \linewidth]{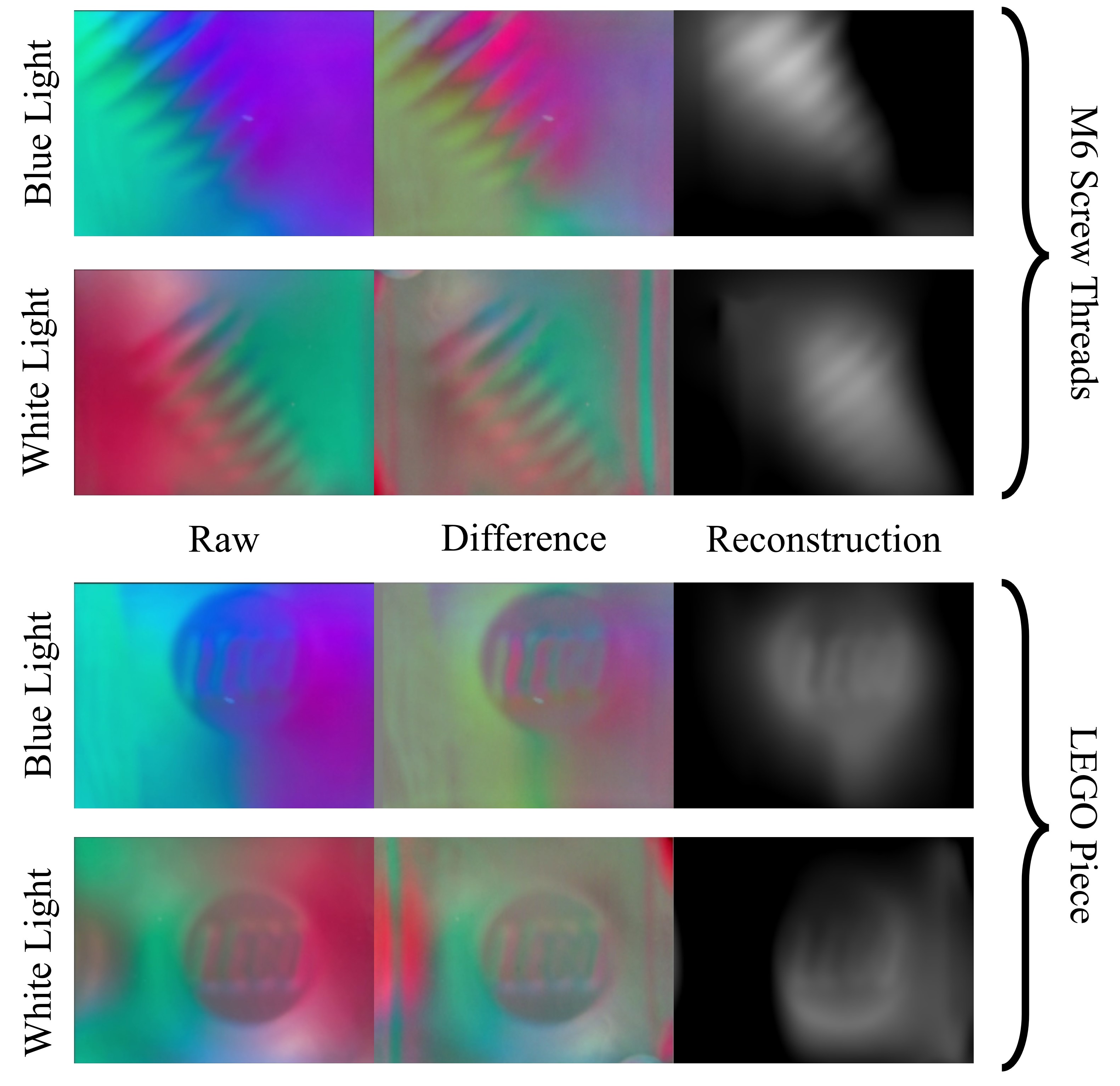}
    \caption{A comparison of the two different illumination systems with an M6 screw and a LEGO piece pressed into the gel surface. Both illumination strategies result in decent 3D reconstruction images. For this paper, we chose to use blue LEDs instead of the warm white LEDs.}
    \label{fig:tactile}
\end{figure}

To get 3D reconstruction of a single tactile image, we take a reference image, an image devoid of any tactile imprints, and subtract it from the tactile image. This difference image is then processed and we get a resulting uncalibrated 3D reconstruction (Fig. \ref{fig:tactile}). We can also use this method for a two-colored light, such as the lighting used in half of the palm design, although the reconstruction will not be as accurate without the addition of a neural network \cite{wedge}.

\section{Experiments and Results}

\subsection{Tactile Sensing}

To test and compare the acuity of the two different tri-color lighting systems, we integrate two ROMEO fingers, one with the blue LED and one with the white LED. We capture videos of the tactile surfaces with objects pressed into them and perform 3D reconstruction on individual objects. 

We found that both the blue LED and white LED illumination methods seem to yield similar 3D reconstruction results (Fig. \ref{fig:tactile}). Both lighting strategies were able to capture details in the individual threads of an M6 screw thread and the words on a LEGO piece connector. However, the white LED had a less even distribution of the RGB colors. As a result, we chose to use blue LEDs and incorporate them into the ROMEO fingers and the passively compliant palm. 

\subsection{Paint Test}

We observe the surface area coverage of a grasped object by implementing a paint test \cite{coolpaint}. The objects we use to test the palm are plastic shape sorting toys with constant profiles (Fisher-Price), and we slather their surfaces with paint. Before the paint dries, we press the side of the object profile into the palm. In total, we use four different shapes: a cylinder, a cube, a plus sign, and a star. We chose these objects for their simple form factors and acknowledge that in the future, varying objects of different shapes and sizes should be used for further rigor. Further, we note that adding fingers to perform direct grasps could vary results based on finger configurations or lengths, so we performed our experiment solely to test the palm by itself.

The different palms we test on are a non-compliant palm, a compliant structure palm, a compliant gel palm, and a palm with both structural and material compliance. All of the palm structures are printed on the Markforged with Onyx material. For the palms with compliant gels, we use a one part mold. 

\begin{figure}[ht!]
    \centering
    \includegraphics[width=1.0 \linewidth]{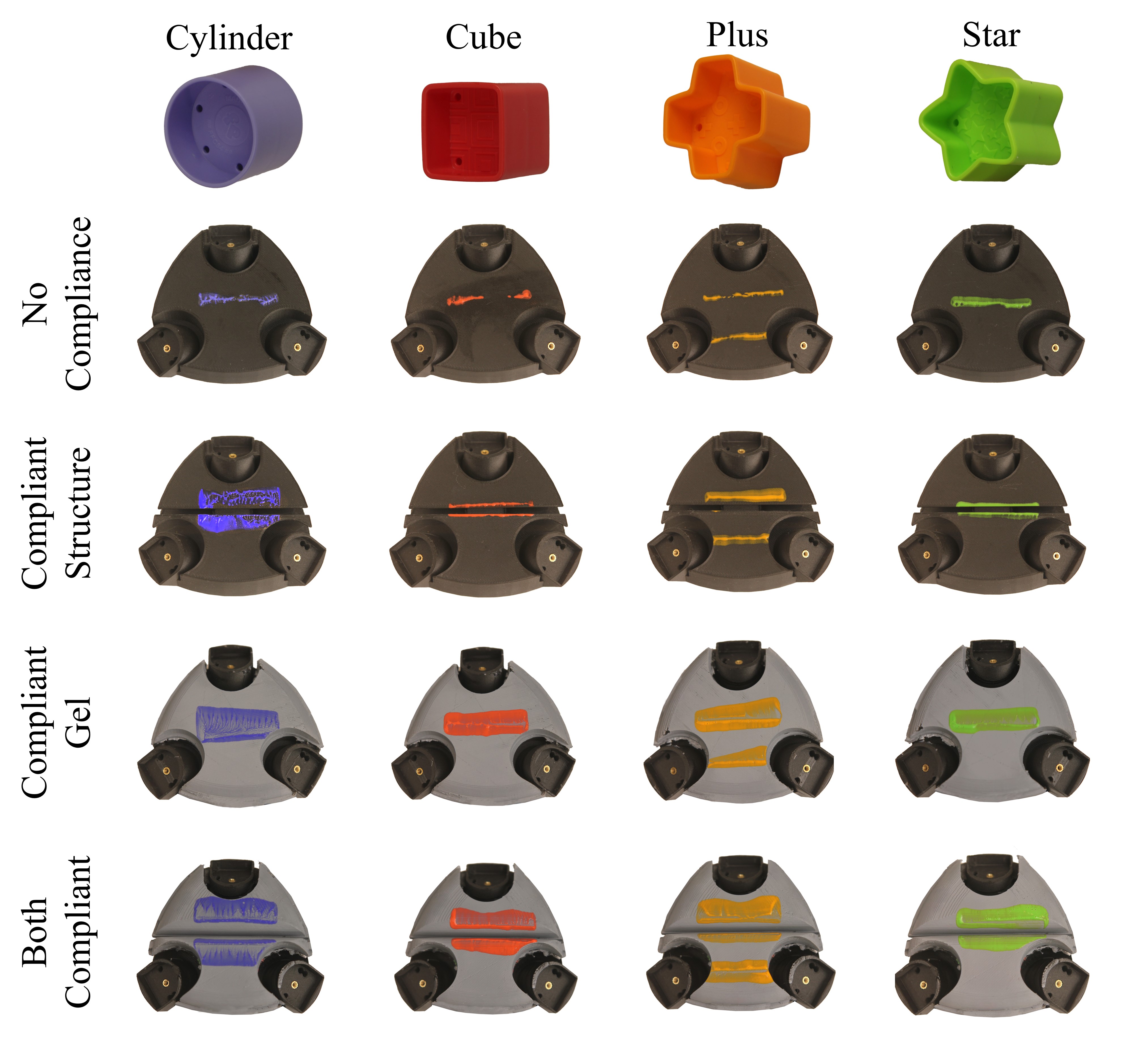}
    \caption{The results of the paint test, where each of the objects were slathered with paint and then pressed into the four different palms. We can visually verify that having both a compliant structure and the compliant gel yields the largest contact area. This result is further corroborated in Table \ref{tab:bleh}.}
    \label{fig:paint}
    \vspace{-10pt}
\end{figure}

Each object is pressed into the palms on a convex corner, if such a corner exists. The same contact point is used on all four palms and the resulting paint presses are shown in Fig. \ref{fig:paint}.

Visually, we can verify that having both structural and material compliance is significantly better than having just one form of compliance or no compliance. Furthermore, we compare the pixel area that each paint imprint covers and verify that having these two compliant systems gives us the most contact area against the palm (Table \ref{tab:bleh}).

\begin{table}[t!]
    \centering
    \caption{Paint Test Comparisons (Pixel Area)}
    \begin{tabular}{ccccc} \hline
    Palm Type & Cylinder & Cube & Plus & Star \\ \hline
    No Compliance & 11380 & 5794 & 12956 & 12920  \\
    Compliant Structure
    & 42347 & 9415 & 24812 & 12281 \\
    Compliant Gel & 44247 & 33963 & 53420 & 33388 \\
    \bf{Both Compliant} & \bf{65135} & \bf{45797} & \bf{69142} & \bf{51228} \\
    \hline \label{tab:bleh} \end{tabular}
    \vspace{-20pt}
    \end{table}

Having a larger area of surface contact means that the grasped object is in a stabler grasp. Thus, our preliminary findings indicate that using both methods of compliance improves the palm functionality. 

\section{Conclusion}

In this paper, we develop a new low-cost illumination system that can be utilized for various soft robotic bodies. We then use this system to design a camera-based tactile finger with an endoskeleton and novel high-resolution tactile palm which incorporates two types of compliance: structural and material. Our design allows the palm to be used in countless different configurations, enabling us to explore a larger space of soft-rigid robotic hand designs with tactile sensors. 

Future work involves designing other types of fingers, which could allow the hand to grasp larger objects, performing more rigorous testing of the palm, and experimenting with the tactile sensing. We also want to be able to streamline and test different hand configurations to optimize for different manipulation tasks.

Overall, we find that implementing both types of compliance in the palm allowed there to be a greater contact area for the grasped object against the palm, creating a more stable grasp. Furthermore, the addition of a palm with tactile sensing capabilities can greatly improve different in-hand tasks that might need tactile sensing such as classification, object reorientation, or the determination of contact geometries. 

\section{Acknowledgements}
Toyota Research Institute, Amazon Science Hub, and the SINTEF BIFROST (RCN313870) project provided funds to support this work. We would also like to express extreme gratitude to Megha H. Tippur for her help in PCB design and soldering. Last but not least, we would like to thank Yuxiang Ma, Laurence Willemet, Leonardo Zamora Ya{\~n}ez, and Jerry Zhang for their helpful tips and insightful conversations. 

\addtolength{\textheight}{-0cm}   



\bibliographystyle{IEEETran}
\bibliography{Ref}

\end{document}